\documentclass[11pt]{article}

\usepackage[preprint]{neurips_2021}

\usepackage{url}            
\usepackage{microtype}      
\usepackage{xcolor}         

\usepackage[utf8]{inputenc} 
\usepackage[T1]{fontenc}    
\usepackage{graphicx}
\usepackage{nicefrac}       

\usepackage{amsmath,amsthm,amssymb,amsfonts}
\usepackage{mathtools}
\usepackage{dsfont}
\usepackage{bm}
\usepackage{bbm} 
\usepackage{relsize}
\usepackage{tikz}
\usetikzlibrary{bayesnet}

\numberwithin{equation}{section}

\usepackage{algorithm}
\usepackage{algorithmicx}
\usepackage{algpseudocode}

\usepackage{booktabs}
\usepackage{caption} 
\usepackage{subcaption}
\usepackage{threeparttable}
\renewcommand{\arraystretch}{1.05}
\usepackage{makecell}

\usepackage{chngcntr}
\usepackage{wrapfig}

\usepackage{multirow}
\usepackage{longtable}
\usepackage{pdflscape}
\usepackage{afterpage}
\usepackage{hyperref}
\usepackage{enumitem}

\usepackage{setspace}
\def\bfB{\mathbf{B}}
\def\bfC{\mathbf{C}}

\def\bfdC{\mathbf{c}}

\def\bfdG{\mathbf{g}}
\def\bfdH{\mathbf{h}}

\def\bfdX{\mathbf{x}}

\def\bfdZ{\mathbf{z}}

\def\bbE{\mathbb{E}}

\def\bbN{\mathbb{N}}

\def\calC{\mathcal{C}}

\def\calE{\mathcal{E}}

\def\calI{\mathcal{I}}
\def\calJ{\mathcal{J}}

\def\indicator{\mathds{1}} 

\newcommand{\DKL}{D_{KL}}

\title{CN-SBM: Categorical Block Modelling For Primary and Residual Copy Number Variation}

\author{%
  Kevin Lam \\
  Department of Statistics\\
  University of British Columbia\\
  \texttt{kevin.lam@stat.ubc.ca}
  \And
    William Daniels, J Maxwell Douglas, Daniel Lai, Samuel Aparicio \\
    Department of Molecular Oncology\\
    BC Cancer Research Centre \\
    \texttt{saparicio@bccrc.ca}
  \And
  Benjamin Bloem-Reddy, Yongjin Park \\
  Department of Statistics\\
  University of British Columbia\\
  \texttt{benbr, ypp@stat.ubc.ca}
}

\begin{document}

\maketitle

\begin{abstract}

Cancer is a genetic disorder whose clonal evolution can be monitored by tracking noisy genome-wide copy number variants. We introduce the Copy Number Stochastic Block Model (CN-SBM), a probabilistic framework that jointly clusters samples and genomic regions based on discrete copy number states using a bipartite categorical block model. Unlike models relying on Gaussian or Poisson assumptions, CN-SBM respects the discrete nature of CNV calls and captures subpopulation-specific patterns through block-wise structure. Using a two-stage approach, CN-SBM decomposes CNV data into primary and residual components, enabling detection of both large-scale chromosomal alterations and finer aberrations. We derive a scalable variational inference algorithm for application to large cohorts and high-resolution data. Benchmarks on simulated and real datasets show improved model fit over existing methods.
Applied to TCGA low-grade glioma data, CN-SBM reveals clinically relevant subtypes and structured residual variation, aiding patient stratification in survival analysis. These results establish CN-SBM as an interpretable, scalable framework for CNV analysis with direct relevance for tumor heterogeneity and prognosis.

\end{abstract}

\section{Introduction}

Heterogeneity in cancer poses a great challenge in cancer genome analysis. Structural alterations, common in tumors, drive key aspects of tumor evolution such as progression, therapy resistance, and subtype diversity.
Copy number variants (CNV) in the single cell DNA sequence~\citep{Laks2019-jz} and tissue-level whole genome sequencing data~\citep{ICGCTCGA_Pan-Cancer_Analysis_of_Whole_Genomes_Consortium2020-yk} provide evidence that cancer genomes are rearranged over the course of clonal evolution. Genome-wide CNV profiles offer a broad overview of these large-scale structural variations, such as aneuploidies, large deletions, and duplications. One way this is accomplished is by partitioning the genome into fixed-size bins (e.g., 500~kb or 1~Mb) and estimating integer copy numbers through read counting and normalization. 

Copy number calling pipelines, such as HMMcopy~\citep{shahIntegratingCopyNumber2006, laiHMMcopyCopyNumber2012} and scAbsolute~\citep{schneiderScAbsoluteMeasuringSinglecell2024}, reduce potential biases (GC and mappability) and generate regularized copy number profiles for downstream analysis. At the single-cell level, it is then possible to investigate cell-to-cell variation that gives rise to subpopulation structures, while sample-to-sample heterogeneity can be characterized using bulk sequencing data, such as that provided by the Cancer Genome Atlas project~\citep{weinsteinCancerGenomeAtlas2013} and even larger data from the ICGC consortium~\citep{ICGCTCGA_Pan-Cancer_Analysis_of_Whole_Genomes_Consortium2020-yk}.

\begin{figure}
    \centering
    \includegraphics[width = 1\linewidth]{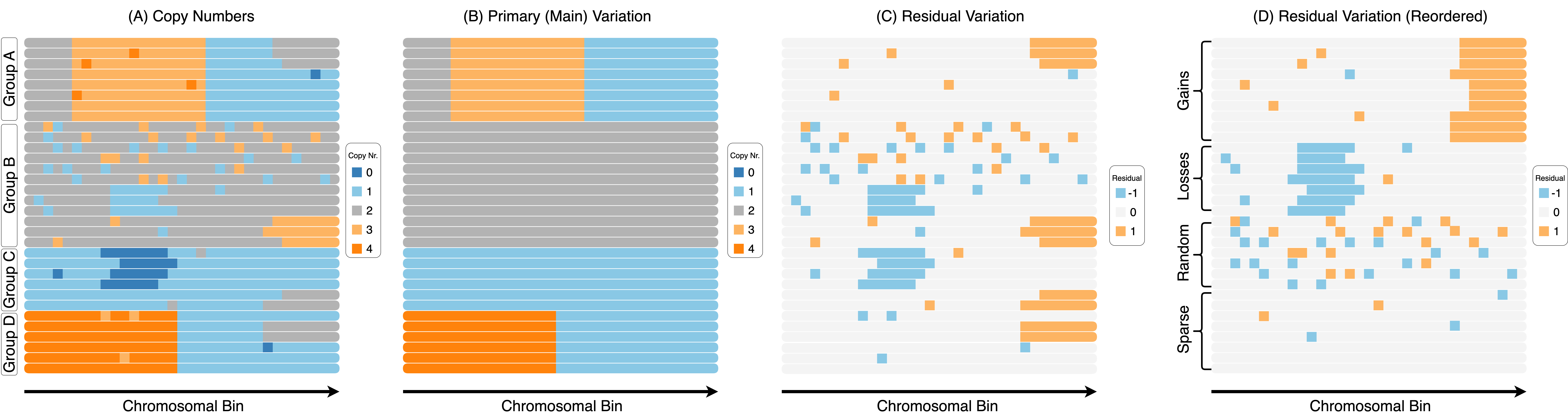}
    \caption{\textbf{Example Separation of Main and Residual Copy Number Variation Across Chromosomal Bins.}
    \textbf{(A)} Raw copy number profiles, visualized per chromosomal bin.
    \textbf{(B)} Main variation pattern capturing the primary copy number trends across samples.
    \textbf{(C)} Residual variation highlighting deviations from the main pattern.
    \textbf{(D)} Residual variation reordered into categories (gains, losses, random, sparse) for clearer interpretation.
    }
    \label{fig: copynumbers}
\end{figure}

CNV profiles show us a stationary view of structural variation, exhibiting characteristic block structures. Here, we explore the possibility that CNV profiles contain richer information, which can be decomposed into primary (main) and residual variation (Fig.~\ref{fig: copynumbers}). Main variation captures large-scale, recurrent alterations that characterize subpopulations, while residual variation reflects finer, cell- and/or sample-specific deviations from these dominant patterns, which may arise from biologically meaningful events such as focal amplifications or deletions. Separating these components enables clearer identification of subpopulation structures while preserving residual signals at the same time~\citep{funnellSinglecellGenomicVariation2022}. This decomposition requires statistical models that can help distinguish structured biological variation from residual variation.


\paragraph{Contributions.} 

An ad-hoc approach to modelling integer copy number states as continuous or count-valued variables using Gaussian or Poisson distributions may be computationally convenient, however, such assumptions can misrepresent the inherently discrete and multimodal nature of CNV data.
To address this, we introduce the \textit{Copy Number Stochastic Block Model (CN-SBM)}, which directly models copy number states using categorical block models~\citep{keribinEstimationSelectionLatent2015}. This avoids restrictive distributional assumptions and captures structured, block-wise variation across subpopulations.

We develop a scalable variational inference scheme for CN-SBM, enabling efficient application to large cohorts and high-resolution genomic data. In contrast to the approach in~\cite{keribinEstimationSelectionLatent2015}, our method updates the variational distributions directly and allows for empty clusters, implicitly enabling automatic model selection for the number of clusters. CN-SBM also lends itself naturally to a two-stage analysis: the first stage captures primary variation, while the second identifies finer-scale residual variations that emerge after accounting for first-stage primary variations. For datasets with approximately 1,000 samples at 500 kb resolution, CN-SBM is computationally efficient, often converging within minutes when accelerated by GPU computing (Fig.~\ref{fig: runtimes}). 

We made our code available at: \href{https://github.com/lamke07/cnsbm2025}{https://github.com/lamke07/cnsbm2025}.

\section{Methods}
\label{section: MNSBM}

\paragraph{Model Setup}

To facilitate modelling, we consider scWGS data in a matrix format, where each row corresponds to a cell, and each column represents the copy number within a fixed-size chromosomal bin (e.g. 500 kilobases (kb)), ensuring consistency in data representation. Let $\bfC = (c_{ij})$ denote the $N \times M$ copy number data matrix, where $c_{ij} \in \bbN_0$ corresponds to the copy number of cell $i$ at bin $j$ across the genome. In the \textit{(bipartite) copy number stochastic block model (CN-SBM)} we propose the use of categorical block models~\citep{keribinEstimationSelectionLatent2015}: in this setup, each cell~$i$ belongs to a \textbf{cell cluster} with up to $K$ components, indexed by $g_i  \in \{1, \ldots, K \}$, and each bin $j$ belongs to a \textbf{bin cluster} with up to $L$ components, indexed by $h_j \in \{1, \ldots, L \}$.

\begin{figure}
    \centering
    \includegraphics[width = 1\linewidth]{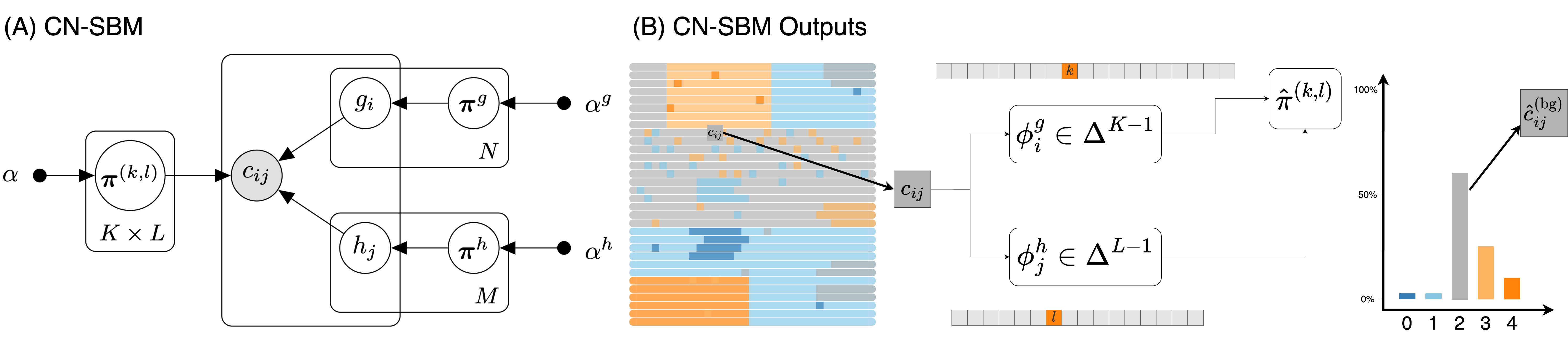}
    \caption{\textbf{Overview of the CN-SBM model for primary variation.}
        \textbf{(A)} Graphical model of the CN-SBM. Observed copy number states $c_{ij}$ are generated based on latent cell ($g_i$) and bin ($h_j$) cluster assignments, with cluster pairs governed by a categorical distribution $\bm{\pi}^{(g_i,h_j)}$.
        \textbf{(B)} Outputs of CN-SBM: for each observed value $c_{ij}$, the \textbf{primary} copy number state $\hat{c}^{(bg)}_{ij}$ is inferred from the most probable cluster pair $(\hat{g}_i,\hat{h}_j)$ under the soft cluster assignments $\phi^g_i, \phi^h_j$ and the corresponding estimated categorical distribution $\hat{\bm{\pi}}^{(\hat{g}_i,\hat{h}_j)}$ over the copy numbers. }
    \label{fig: graphical_model_sbm}
\end{figure}

The observed copy number $c_{ij}$ is modelled as a categorical variable whose distribution depends only on the latent cluster pair $(g_i, h_j)$ (Fig.~\ref{fig: graphical_model_sbm}). Copy number values are assumed to lie in a fixed, discrete set $\calC = \{ 0, 1, \ldots, \geq 11\}$ with $|\calC| = n_\text{cat}$, where highly amplified values are grouped into a final category. The generative process is:
\begin{enumerate}
    \item For each cell $i \in \{1, \ldots, N\}$, draw cell cluster assignment $g_i \sim \text{Cat}(\bm{\pi}^g)$.
    \item For each bin $j \in \{1, \ldots, M\}$, draw bin cluster assignment $h_j \sim \text{Cat}(\bm{\pi}^h)$ .
    \item For each pair $(i,j)$, draw copy number $c_{ij} \sim \text{Cat}\left(\bm{\pi}^{(g_i,h_j)}\right)$, where $\bm{\pi}_{k,l}$ is the categorical distribution over copy number values for the cluster pair $(k, l)$.
\end{enumerate}
We place Dirichlet priors over all categorical distributions, that is, $\bm{\pi}^g \sim \text{Dir}_{[K]}(\alpha^g)$ and $\bm{\pi}^h \sim \text{Dir}_{[L]}(\alpha^h)$ for the possible cell and bin cluster assignments, respectively, and $\bm{\pi}^{(k,l)} \sim \text{Dir}_{[n_\text{cat}]}(\alpha)$, for each of the $K\cdot L$ individual block (cluster) distributions, $1\leq k \leq K, 1 \leq l \leq L$.
Assuming independence among cluster assignments and copy number observations, the joint distribution factorizes as:
\begin{equation*}
\begin{split}
    p(\bfC, \bfdG, \bfdH, \bm{\pi}^g, \bm{\pi}^h, \{\bm{\pi}^{(k,l)}\}) &=
    \underbrace{\prod_{i=1}^N \prod_{j=1}^M \text{Cat}(c_{ij}; \bm{\pi}^{(g_i,h_j)})}_{\text{data model}}
     \cdot \underbrace{\prod_{i=1}^N \text{Cat}(g_i; \bm{\pi}^g)}_{\text{cell clusters}}
    \cdot \underbrace{\prod_{j=1}^M \text{Cat}(h_j; \bm{\pi}^h)}_{\text{bin clusters}} \\
    &\quad \cdot \underbrace{\prod_{k=1}^K \prod_{l=1}^L \left( \text{Dir}_{[n_\text{cat}]}(\bm{\pi}^{(k,l)}; \alpha)\right)}_{\text{block distribution prior}} 
    \cdot \underbrace{\text{Dir}_{[K]}(\bm{\pi}^g; \alpha^g)}_{\text{cell cluster prior}}
    \cdot \underbrace{\text{Dir}_{[L]}(\bm{\pi}^h; \alpha^h)}_{\text{bin cluster prior}}.
\end{split}
\end{equation*}
Here, $\{\bm{\pi}^{(k,l)}\} \coloneqq \{\bm{\pi}^{(k,l)}\}_{k \in [K], l \in [L]}$ denotes the collection of all $K\cdot L$  block-specific probability vectors. Using this factorization, the complete conditional distributions for the latent variables are:
\begin{equation}
\begin{split}
    p(g_i = k| \bfdH, \bm{\pi}^g, \{\bm{\pi}^{(k,l)}\}, \bfdC_{i,:}) &\propto \exp\left(\sum_{j=1}^M \sum_{l=1}^L \indicator(h_j = l) \log \pi^{(k,l)}_{c_{ij}} + \log \pi^g_k  \right) \\
    p(h_j = l| \bfdG, \bm{\pi}^h, \{\bm{\pi}^{(k,l)}\}, \bfdC_{:,j}) &\propto \exp\left(\sum_{i=1}^N \sum_{k=1}^K \indicator(g_i = k) \log \pi^{(k,l)}_{c_{ij}} + \log \pi^h_l \right) \\
    p (\bm{\pi}^g | \bfdG) &= \text{Dir}_{[K]}\left(\bm{\pi}^g; \alpha^g + \sum_{i=1}^N \indicator(g_i = \cdot)\right) \\
    p (\bm{\pi}^h | \bfdH) &= \text{Dir}_{[L]}\left(\bm{\pi}^h; \alpha^h + \sum_{j=1}^M \indicator(h_j = \cdot)\right) \\
    p (\bm{\pi}^{(k,l)} | \bfdG, \bfdH, \bfC) &= \text{Dir}_{[n_\text{cat}]}\left(\bm{\pi}^{(k,l)}; \alpha + \sum_{i=1}^N\sum_{j=1}^M \indicator(g_i = k)\indicator(h_j = l) \indicator(c_{ij} = \cdot)\right),
    \label{eq: mnsbm_complete_conditionals}
\end{split}
\end{equation}
where we have leveraged the conjugacy between the Dirichlet and categorical distributions for the latent probability vectors $\bm{\pi}^g$, $\bm{\pi}^h$, $\{\bm{\pi}^{(k,l)}\}$. The `$\cdot$' symbol in the indicator functions denotes the appropriate index over the relevant latent variable dimension, allowing for a concise representation of cluster-specific counts.


\paragraph{Variational Inference}

The CN-SBM assumes a latent variable model setup of the form $p(\bfC, \bfdZ) = p(\bfC \vert \bfdZ) p(\bfdZ)$, where $\bfdZ = (\bfdG, \bfdH, \bm{\pi}^g, \bm{\pi}^h, \{\bm{\pi}^{(k,l)}\})$ contains all latent variables.
We adopt a mean-field variational approximation~\citep{ghahramaniPropagationAlgorithmsVariational2000} of the posterior that factorizes over the latent variables as:
\begin{equation*}
\begin{split}
    q(\bfdG, \bfdH, \bm{\pi}^g, \bm{\pi}^h, \{\bm{\pi}^{(k,l)}\}) 
    &= \prod_{i=1}^N q(g_i | \bm{\phi}^g) \cdot \prod_{j=1}^M q(h_j | \bm{\phi}^h)
     \cdot \prod_{k=1}^K \prod_{l=1}^L \left[ q(\bm{\pi}^{(k,l)} | \bm{\gamma}^{(k,l)}) \right] \cdot q(\bm{\pi}^g | \bm{\gamma}^g) \cdot q(\bm{\pi}^h | \bm{\gamma}^h),
\end{split}
\end{equation*}
where $q(g_i = k| \bm{\phi}^g) = \phi^g_{ik}$ and $q(h_j = l| \bm{\phi}^h) = \phi^h_{jl}$ are variational distributions corresponding to soft cluster assignments for the cells and bins. The variational distributions over the Dirichlet parameters are modeled as $\bm{\pi}^g \sim \text{Dir}_{[K]}\left(\bm{\gamma}^g \right)$, $\bm{\pi}^h \sim \text{Dir}_{[L]}\left(\bm{\gamma}^h \right)$, and $\bm{\pi}^{(k,l)} \sim \text{Dir}_{[n_\text{cat}]}\left(\bm{\gamma}^{(k,l)} \right)$ for each block $(k,l)$.
The evidence lower bound (ELBO) on the log-marginal likelihood $\log p(\bfC)$ is:
\begin{equation*}
    \log p(\bfC)
    \geq \bbE_{q}[\log p(\bfC | \bfdG, \bfdH, \{\bm{\pi}^{(k,l)}\})] -\DKL (q(\bfdG, \bfdH, \bm{\pi}^g, \bm{\pi}^h, \{\bm{\pi}^{(k,l)}\}) || p(\bfdG, \bfdH, \bm{\pi}^g, \bm{\pi}^h, \{\bm{\pi}^{(k,l)}\})),
\end{equation*}
which can be computed analytically in the CN-SBM setup (Appendix~\ref{appendix: sbm_details}). We aim to maximize the ELBO using \textbf{Coordinate Ascent Variational Inference (CAVI)}~\citep{bleiVariationalInferenceDirichlet2006, bishopPatternRecognitionMachine2006, bleiVariationalInferenceReview2017}, an iterative optimization algorithm for Bayesian inference. For a factorized variational distribution $q(\bfdZ) = \prod_{i=1}^m q_i(\bfdZ_i)$, the optimal update for each $q_j(\bfdZ_j)$ while holding the remaining variables $\bfdZ_i, i \neq j$, fixed is
\begin{equation}
    q_j(\bfdZ_j) \propto \exp \left(\bbE_{i\neq j} [\log p(\bfdZ_j | \bfdZ_{-j}, \bfdX)]\right),
\label{eq:cavi_update}
\end{equation}
where the expectation is taken with respect to the variational distributions over all other latent variables $\bfdZ_{-j}$. This update has a closed-form expression requiring only natural parameter updates when the complete conditionals $p(\bfdZ_j | \bfdZ_{-j}, \bfdX)$ belong to an exponential family, which is the case in the CN-SBM framework. In particular, the variational update steps are analytically derived by computing expectations over the conditionals presented in eq.~\eqref{eq: mnsbm_complete_conditionals}.
The full variational inference scheme is outlined in algorithm~\ref{algo:sbm_cavi}. In this context, $\psi$ is the digamma function, which provides the expectation of a log-transformed Dirichlet variable, given by the form $\bbE[\log \pi_{k'}] = \psi(\alpha_{k'}) - \psi\left(\sum_{k} \alpha_k\right)$. This expectation quantifies the relative importance of category $k'$ within the total concentration of the Dirichlet distribution. These updates are guaranteed to converge to a local optimum of the ELBO~\citep{boydConvexOptimization2004, bishopPatternRecognitionMachine2006}.
\begin{algorithm}[!htb]
\small
\caption{Variational inference for the CN-SBM}
\begin{algorithmic}
\Require Copy number matrix $\bfC = (c_{ij})$.
\State Initialize of local parameters $\bm{\phi}^g, \bm{\phi}^h$ and global parameters $\bm{\gamma}^g, \bm{\gamma}^h, \{\bm{\gamma}^{(k,l)}\}$.
\While{ELBO not converged}
    \State \textbf{Local updates for each row and column}
    \begin{itemize}
        \item for row $i = 1, \ldots N$ and row clusters $k = 1, \ldots, K$
        \begin{itemize}
            \item $q(g_i = k) \propto \exp\left(\sum_{j=1}^M \sum_{l=1}^L \phi^h_{jl} \cdot \left(\psi(\gamma^{(k,l)}_{c_{ij}}) - \psi\left(\sum_{c \in \calC} \gamma^{(k,l)}_c \right) \right)  + \psi(\gamma^g_k) - \psi\left(\sum_{k=1}^K \gamma^g_k\right)  \right)$
        \end{itemize}
        \item for column $j = 1, \ldots M$ and column clusters $l = 1, \ldots, L$
        \begin{itemize}
            \item $q(h_j = l) \propto \exp\left(\sum_{i=1}^N \sum_{k=1}^K \phi^g_{ik} \cdot \left(\psi(\gamma^{(k,l)}_{c_{ij}}) - \psi\left(\sum_{c \in \calC} \gamma^{(k,l)}_c \right) \right)  + \psi(\gamma^h_l) - \psi\left(\sum_{l=1}^L \gamma^h_l\right)  \right)$
        \end{itemize}
    \end{itemize}
    \State \textbf{Global updates for cluster and cluster proportions}
    \begin{itemize}
        \item $\bm{\gamma}^g = \alpha^g + \sum_{i=1}^N \phi^g_{i, \cdot}$
        \item $\bm{\gamma}^h = \alpha^h + \sum_{j=1}^N \phi^h_{j, \cdot}$
        \item $\bm{\gamma}^{(k,l)} = \alpha + \sum_{i=1}^N\sum_{j=1}^M  \phi^g_{ik} \phi^h_{jl} \indicator(c_{ij}=\cdot), \quad \quad 1 \leq k \leq K, 1\leq l \leq L.$
    \end{itemize}
\EndWhile
\end{algorithmic}
\label{algo:sbm_cavi}
\end{algorithm}

The local updates for cell cluster assignments $g_i$ and bin cluster assignments $h_j$ can be viewed as soft assignments over $K$ and $L$ clusters, respectively, where each cluster’s relevance is determined by expected log-probabilities of the observed data categories. These expectations are computed as weighted averages over the columns (for cells) or rows (for bins), using the current corresponding soft responsibilities $\phi^h_{jl}$ and $\phi^g_{ik}$. The prior expectations $\mathbb{E}_q[\log \pi^g_k]$ and $\mathbb{E}_q[\log \pi^h_l]$ act as regularizers, reflecting the influence of Dirichlet priors. Global parameters $\bm{\gamma}^g$, $\bm{\gamma}^h$, and $\bm{\gamma}^{(k,l)}$ summarize the local assignments by aggregating responsibilities across cells, bins, and block interactions, with each update incorporating the corresponding prior parameters $\alpha^g, \alpha^h$ and $\alpha$, respectively.


\paragraph{Missing Data}

To handle missing values, we extended the CAVI algorithm to optimize the ELBO over the full data-generating process, bounding $\log p(\bfC_\text{obs}, \bfC_\text{mis})$, rather than just $\log p(\bfC_\text{obs})$. This encourages the latent structure to reflect the entire dataset, rather than only the observed portion. Let $\bfB = (b_{ij})$ denote the binary missingness matrix, where $b_{ij} = 1$ if $c_{ij}$ is observed. We introduce importance weights $w_{ij}$ to adjust each observation’s influence, applying them as multiplicative factors in the variational updates in Algorithm~\ref{algo:sbm_cavi} and ELBO computations (Appendix~\ref{section: missing}). For instance, setting $w_{ij} = b_{ij}$ ignores missing entries, while $w_{ij} = b_{ij} / \hat{\zeta}_{ij}$ incorporates inverse-propensity weighting via an estimated missingness model $\hat{\zeta}_{ij}$, e.g. from logistic regression or frequency-based heuristics.

\paragraph{Initialization and Model Refinement}

Since CAVI guarantees only local convergence, the choice of initialization of variational parameters affects the quality of the final solution, with the converged ELBO often varying substantially across runs~\citep{bleiVariationalInferenceReview2017}. To improve robustness and convergence, we support several initialization schemes for the variational parameters. These include random sampling from the prior and clustering-informed initialization such as k-means~\citep{macqueenMethodsClassificationAnalysis1967, lloydLeastSquaresQuantization1982} and spectral clustering~\citep{ngSpectralClusteringAnalysis2001} applied independently to rows and columns. For joint initialization, we also incorporate spectral biclustering~\citep{klugerSpectralBiclusteringMicroarray2003}, which empirically yielded the best performance in our experiments.

Given a fitted model, local exploration strategies can be applied that search for improved optima by splitting or merging clusters based on the posterior MAP hard assignments $\hat{\bfdG}_\text{MAP}, \hat{\bfdH}_\text{MAP}$. These splits can be performed by reassigning labels within existing sub-clusters either via standard clustering methods or by assigning new labels according to the categorical mode. To correct for over-segmentation, clusters can be merged when redundancy is detected, such as nearly identical dominant categories or when cluster sizes fall below a certain threshold, based on the number of cells (rows) or bins (columns). Throughout this process, model selection can be guided by improvements in the ELBO or by penalized criteria such as the integrated completed likelihood (ICL)~\citep{biernackiAssessingMixtureModel2000, comeModelSelectionClustering2015, bar-henBlockModelsGeneralized2022}, which account for both model fit and complexity.

\section{Experiments}
\label{subsection: sbm_initial_results}

\paragraph{Datasets}

We consider three datasets for our experiments, all requiring a copy number matrix with rows as cells or patient samples and columns as genomic bins. Since the CN-SBM accepts categorical inputs, it can accommodate both single-cell and bulk copy number data. First, we generated 2,500 single-cell copy number profiles across eight clones using the \texttt{CNAsim} simulator~\citep{weinerCNAsimImprovedSimulation2023}. This simulation includes chromosome-arm and whole-chromosome-level copy number alterations, along with configurable noise to control segment lengths and per-bin copy number jitter. The resulting profiles were aggregated into non-overlapping 500 kb genomic bins. Next, we incorporated single-cell data from the study by~\citet{funnellSinglecellGenomicVariation2022}, which includes DLP+ sequencing of patient-derived xenografts (PDX) from various tumours collected over multiple time points spanning several years.
Copy number calling was performed using HMMCopy~\citep{shahIntegratingCopyNumber2006, laiHMMcopyCopyNumber2012} at 500kb resolution, yielding total copy number profiles. Specifically, we considered the OV2295 ($N=1084$), SA1096 ($N=802$) and SA535 ($N=1801$) cell lines.

Lastly, we also evaluated our method on bulk sequencing data from The Cancer Genome Atlas (TCGA) project~\citep{weinsteinCancerGenomeAtlas2013}, where each matrix row corresponds to a patient tumour sample. Copy number states were previously inferred using the ASCAT algorithm~\citep{vanlooAllelespecificCopyNumber2010a}, which estimates allele-specific copy numbers. We rebinned the segmented data into 500 kb bins to match the resolution used in our single-cell analyses and computed the total copy number as the sum of major and minor alleles. Unlike single-cell data, this dataset reflects inter-patient population-level copy number diversity rather than intra-tumour heterogeneity. For benchmarking, we selected breast cancer ($N=998$), ovarian cancer ($N=532$), and low-grade glioma ($N=490$).
Some datasets have missing values in the copy number matrix, which can be due to low sequencing coverage, high-variability regions (e.g., mutation hotspots), or technical artifacts like alignment errors and GC-content bias. Since standard clustering algorithms, including those in our study, do not handle missing data, we impute missing entries by sampling from the empirical marginal distribution of observed copy number states across five seeds.


\paragraph{Experiment Settings}

We evaluate two experimental settings. In the first, models are trained on the fully imputed dataset to assess clustering performance under complete-data assumptions. In the second, we simulate additional missingness by randomly withholding approximately 1\% of observed entries, which are excluded during training and used as a held-out set to evaluate predictive accuracy. Specifically, we assess how well the fitted categorical stochastic block model infers the unseen values, based on the posterior over latent assignments and block parameters.
We benchmarked our method against several co-clustering approaches, including the Poisson bipartite stochastic block model (PoissonSBM)~\citep{bar-henBlockModelsGeneralized2022, chiquetSbmStochasticBlockmodels2024}, the categorical latent block model (Blockcluster)~\citep{keribinEstimationSelectionLatent2015, bhatiaBlockclusterPackageModelBased2017}, k-means clustering (KMeans), and spectral biclustering with log-transformed or bi-stochastic input (SpecBi). All methods can produce discrete partitions over the rows (cells/samples) and columns (genomic bins), allowing direct comparison. We excluded biclustering methods that allow overlapping cluster memberships, such as the FABIA~\citep{hochreiterFABIAFactorAnalysis2010} and Cheng and Church algorithm~\citep{chengBiclusteringExpressionData2000}.
\begin{table}[ht]
\centering
\small
\vspace{1mm}
\renewcommand{\arraystretch}{1.1}
\setlength{\tabcolsep}{2.5pt}
\begin{tabular}{|c|l|c|ccc|ccc|}
\hline
 & \textbf{Method} & \textbf{CNAsim} & \multicolumn{3}{c|}{\textbf{TCGA}} & \multicolumn{3}{c|}{\textbf{Funnell}} \\
\cline{3-9}
 &  & \makecell{Simulated \\ $(N=2500)$} & \makecell{BRCA \\ $(N=998)$} & \makecell{OV \\ $(N=532)$} & \makecell{LGG \\ $(N=490)$} & \makecell{OV2295 \\ $(N=1084)$} & \makecell{SA1096 \\ $(N=802)$} & \makecell{SA535 \\ $(N=1801)$} \\
\hline\hline
\multirow{6}{*}{\makebox[6pt][c]{\rotatebox[origin=c]{90}{\textbf{LL} $\times 10^3$ ($\uparrow$)}}}
& SpecBi (bist)         & -89.1 ± 0.3      & -75.3 ± 0.1  & -45.5 ± 0.3  & -19.0 ± 0.2    & -54.7 ± 1.0    & -49.0 ± 0.4  & -107.2 ± 1.7 \\
 & SpecBi (log)          & -88.1 ± 0.6      & -74.7 ± 0.2  & -44.4 ± 0.1  & -18.8 ± 0.2    & -55.5 ± 1.0    & -48.2 ± 0.8  & -97.5 ± 1.1 \\
 & KMeans                & -41.0 ± 0.6      & -60.1 ± 0.2  & -37.6 ± 0.1  & \textbf{-11.0 ± 0.1}    & -37.7 ± 0.6    & -27.3 ± 0.4  & -43.9 ± 1.0 \\
 & PoissonSBM            & -41.6 ± 3.0      & -115.8 ± 18.0 & -59.7 ± 2.9  & -125.5 ± 32.8  & -60.4 ± 8.2    & -46.0 ± 8.0  & -44.3 ± 3.3 \\
 & Blockcluster          & -38.1 ± 0.4      & \textbf{-57.0 ± 0.2}  & -37.0 ± 0.3  & -11.5 ± 0.3    & -39.0 ± 0.8    & -27.7 ± 0.4  & -46.9 ± 1.7 \\
\cline{2-9}
 & CN-SBM (ref.)         & \textbf{-37.6 ± 0.3} & -57.1 ± 0.1 & \textbf{-36.8 ± 0.1} & -11.3 ± 0.2 & \textbf{-36.3 ± 1.2} & \textbf{-26.5 ± 0.5} & \textbf{-42.5 ± 0.4} \\
\hline\hline
\multirow{6}{*}{\makebox[6pt][c]{\rotatebox[origin=c]{90}{\textbf{ICL} $\times 10^6$ ($\uparrow$)}}}
  & SpecBi (bist)         & -8.89 ± 0.02  & -7.52 ± 0.03 & -4.58 ± 0.02 & -1.92 ± 0.02 & -5.62 ± 0.07 & -4.99 ± 0.10 & -11.27 ± 0.61 \\
 & SpecBi (log)          & -8.82 ± 0.07  & -7.52 ± 0.05 & -4.47 ± 0.01 & -1.93 ± 0.02 & -5.58 ± 0.13 & -4.95 ± 0.06 & -10.08 ± 0.40 \\
 & KMeans                & -4.04 ± 0.09  & -6.02 ± 0.03 & -3.81 ± 0.01 & -1.13 ± 0.01 & -3.72 ± 0.03 & -2.76 ± 0.02 & -4.42 ± 0.05 \\
 & PoissonSBM            & -5.31 ± 0.00  & -12.77 ± 1.26 & -6.06 ± 0.21 & -15.97 ± 1.12 & -5.08 ± 0.00 & -5.28 ± 0.00 & -4.35 ± 0.00 \\
 & Blockcluster          & -3.78 ± 0.06 & -5.72 ± 0.01 & -3.74 ± 0.01 & -1.21 ± 0.01 & -4.11 ± 0.11 & -- & -4.68 ± 0.04 \\
\cline{2-9}
 & CN-SBM (ref.)         & \textbf{-3.58 ± 0.02} & \textbf{-5.62 ± 0.01} & \textbf{-3.70 ± 0.01} & \textbf{-1.12 ± 0.01} & \textbf{-3.54 ± 0.10} & \textbf{-2.67 ± 0.02} & \textbf{-4.20 ± 0.09} \\
\hline
\end{tabular}
\caption{Held‐out log‐likelihood (top) and ICL (bottom) across simulated and real datasets.
Values were scaled by $10^3$ for log-likelihood and $10^6$ for ICL. Dashes indicate missing values due to failed runs.
}
\label{tab:combined_loglik_ICL}
\end{table}

We fixed the maximum number of row and column clusters to $K = 10, L = 30$ for the CNAsim and TCGA datasets, and $K = 15, L = 30$ for the Funnell datasets, based on prior cluster expectations and practical run-time constraints. For example, PoissonSBM uses top-down hierarchical agglomeration with repeated model fitting to optimize the ICL, making model selection computationally prohibitive. KMeans and spectral biclustering do not provide probabilistic outputs. As such, we estimated empirical categorical distributions within each block (row-column cluster pair) using observed frequency counts. The block distributions were estimated as $\hat{\pi}^{(k, l)}_c = (n^{(k, l)}_c) / (\sum_{c'}n^{(k, l)}_{c'})$, where $n^{(k, l)}_c$ is the count of copy number state $c$ in block $(k, l)$.
These distributions were used to compute ICL and held-out log-likelihood scores.
For PoissonSBM and Blockcluster, we also used empirical block distributions, as the inferred probability distributions from their implementations generally yielded poorer results. For PoissonSBM, the assumed data distribution struggles to assign high probability to individual counts, while the block probabilities inferred by Blockcluster sometimes produced extreme ICL and log-likelihood values, distorting average performance metrics (Appendix~\ref{section: add_benchmark_details} and Table~\ref{tab:combined_loglik_ICL_alt}).


\paragraph{Model Fit}

Table~\ref{tab:combined_loglik_ICL} provides an overview of how the CN-SBM performs against the other baselines. Three complementary criteria are reported: (i) held-out log-likelihood, (ii) integrated completed-likelihood (ICL), which penalises overly complex latent structures while rewarding good data fit, and (iii) weighted average normalized entropy to assess cluster purity (Supplementary Table~\ref{tab:combined_entropy}). The model \textit{CN-SBM (ref.)} explores the partition space using posterior parameters from a single CN-SBM run, similarly to PoissonSBM.

Model quality is evaluated using held-out log-likelihood, measuring how well the predicted distribution aligns with unseen data, and the ICL, which assesses overall model fit on the full training data. The CN-SBM achieves the highest held-out log-likelihood across most datasets, and performance margins generally increased with the size of the data. 
The large performance gap between PoissonSBM and other methods on the TCGA datasets and OV2295 is due to how PoissonSBM handles cluster specification. When given $K$ row and $L$ column clusters, it searches over a total of $K + L$ clusters without fixing the row and column counts. As a result, the final model often assigns more row or column clusters than specified, for example, averaging 25 row clusters in BRCA despite $K = 10$, and similarly for OV and LGG. This may be due to the wide range of copy number variation that cannot be reflected by a unimodal count distribution such as the Poisson distribution. The ICL includes a complexity penalty and thus favours simpler models when performance is comparable. The CN-SBM yields the highest ICL across all datasets (Table~\ref{tab:combined_loglik_ICL}), indicating that its inferred block structure and block distributions effectively capture underlying data patterns best.

As a second order assessment, we also computed the average normalized entropy weighted by subcluster size to analyze the within-cluster purity of the inferred sub-clusters, weighted by sub-cluster sizes. Low entropy indicates clear separation between primary copy number variation and residual variation, enabling the identification of distinct groups without confounding noise and reflecting the model’s ability to capture major sources of variation. We report the results in Supplementary Table~\ref{tab:combined_entropy}: the CN-SBM achieves the lowest weighted entropy across most datasets, significantly outperforming spectral biclustering and showing similar performance to PoissonSBM and Blockcluster.

Standard deviations across five runs remain comparable with the baseline methods for the CN-SBM, highlighting its robustness despite variation in cell/sample counts and copy number patterns. Blockcluster often failed to converge (see SA1096 and Table~\ref{tab:model_runs}). For the PoissonSBM, the Funnell datasets (OV2295, SA1096, SA535) and CNAsim are fully observed, so no imputation was needed, which eliminates a potential source of variability. As such, PoissonSBM showed almost no variance on these datasets due to its agglomerative fitting process, which consistently converged to the same cluster partitioning across random initializations.


\section{Primary and Residual Variation in Low-Grade Glioma}

\begin{figure}[!htb]
    \centering
    \includegraphics[width = 1\linewidth]{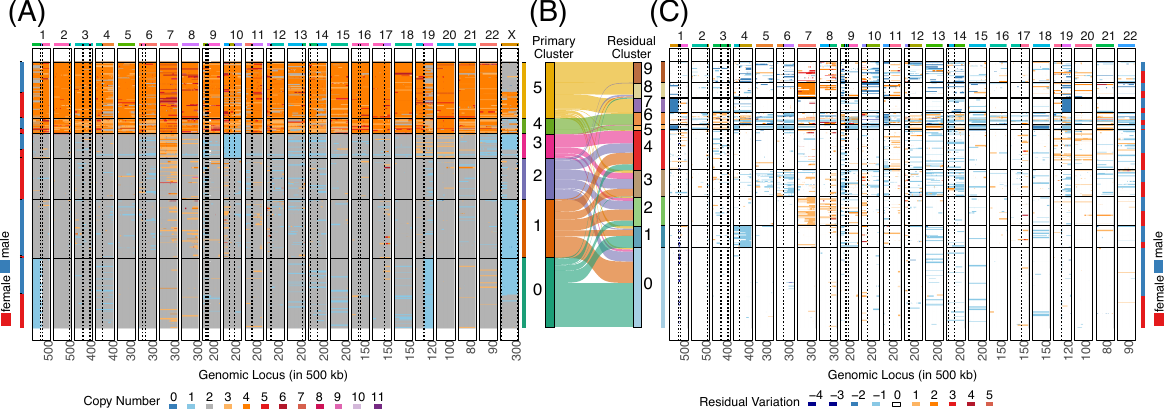}
    \caption{
        \textbf{Main and residual variation in TCGA lower-grade glioma ($N=484$)}.
        \textbf{(A)} CN-SBM fitted to the LGG copy number profiles in 500kb resolution, with samples and bins coloured according to their cluster.
        \textbf{(B)} Alluvial diagram linking sample cluster assignment in main variation (A)  vs. residual variation (C).
        \textbf{(C)} Residual copy-number landscape after subtraction of main variation obtained from the CN-SBM. Samples have been reordered according to the clustering obtained from a second categorical SBM fitted to this matrix, identifying ten finer \textit{residual} clusters.
    }
    \label{fig: LGG_results}
\end{figure}

We analyze somatic copy number alterations from the TCGA low-grade glioma (LGG) dataset, comprising $N = 490$ samples. Copy number states are thresholded at 11, with high or extreme amplifications categorized as 11 to reduce sparsity. To capture the main variation, we fit the CN-SBM and assign hard clusters. Within each co-cluster (block), we summarize copy number patterns by selecting the copy number with the highest posterior Dirichlet-mean (Fig.~\ref{fig: graphical_model_sbm}B), forming a \textit{main variation} matrix of representative values per sample-bin pair. Subtracting this matrix from the original copy number profiles yields a residual matrix, with negative values indicating underrepresentation. We re-encode these deviations categorically and apply the CN-SBM again to capture finer structure. As before, hard cluster assignments are obtained by computing the posterior argmax assignment. This two-stage approach decomposes copy number variation into main and residual components using categorical block models. The \texttt{X} chromosome is excluded in the second stage to avoid sex-related confounding.

Application of the CN-SBM identified ten \textit{main (variation)} clusters within the samples. Four of these, comprising just six outlier samples with profiles dissimilar form the other main patterns, were excluded to focus on six dominant clusters ($N = 484$) that capture the primary structure of copy number variation (Fig.~\ref{fig: LGG_results}A). Reordering the copy number matrix by cluster assignments reveals distinct genomic signatures. Clusters 0-3 display arm-level and whole-chromosome alterations involving chromosomes 1, 7, 10, 19, and X. In contrast, clusters 4 and 5 are marked by widespread amplification, with modal states near four, consistent with whole-genome doubling events.

Having applied the CN-SBM to the residual matrix, the second-stage analysis uncovered structured patterns that were not apparent in the primary clustering. As illustrated in the Sankey diagram (Fig.~\ref{fig: LGG_results}B), samples from the same primary cluster often split into distinct residual clusters, indicating shared patterns of deviation within otherwise similar profiles (Fig.~\ref{fig: LGG_results}C).
Residual cluster 0 shows minimal deviation, suggesting strong alignment with the dominant structure. In contrast, other residual clusters reveal specific alterations: cluster 1 shows chromosome 4 loss; cluster 2 exhibits gains on chromosomes 7 and 8; cluster 3 is characterized by widespread deletions affecting chromosomes 5, 6, 9, 13, 14, and 19. Additional residual clusters 5, 7, and 8 display complex patterns involving chromosomes 1, 7, and 19. These may highlight biologically relevant variation that emerge after characterizing the primary variations.

\begin{figure}[!htb]
    \centering
    \includegraphics[width = 1\linewidth]{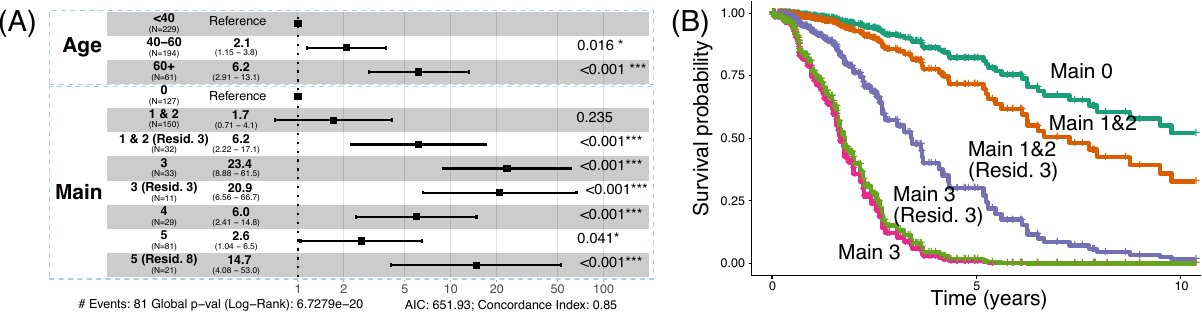}
      \caption{
        \textbf{(A)} Forest plot for Cox proportional hazards model evaluating impact of age groups and \textit{main} cluster membership on patient survival. Distinct copy-number profiles segregate patients into different survival trajectories (log-rank $p < 0.001$).
        \textbf{(B)} Kaplan-Meier survival curves for patient profiles with 40-60 age group and varying clusters.
      }
    \label{fig: LGG_results_surv}
\end{figure}

To evaluate the prognostic relevance of the CN-SBM-inferred primary (main) and residual variation, we performed Cox proportional hazards modelling using patient survival data. The cohort comprised 213 female patients (44\%), with ages ranging from 14 to 87 years (median 41.0 years, IQR 32.8-52.8). Age was discretized into three categories: less than 40, 40-60, and 60+ years. In an initial model excluding CN-SBM clusters, sex was not significantly associated with survival, which is consistent with its balanced distribution across main copy number clusters (clusters 0, 3, 4, and 5; Fig.~\ref{fig: LGG_results}A). As such, the baseline model included only age group as a covariate and achieved a concordance index (C-index) of 0.741 (SE = 0.031).

Through model selection (Appendix~\ref{appendix: cox_model_selection}), we incorporated both the main and part of the residual CN-SBM cluster assignments. This substantially improved model analytics, yielding a C-index of 0.855 (SE = 0.019) (Fig.~\ref{fig: LGG_results_surv}A). This highlights the strong predictive and explanatory value of the discovered genomic subtypes. Most main clusters were significantly associated with survival: clusters 1 and 2 were combined (\textit{1 \& 2}) due to minimal differences, primarily limited to \texttt{X} chromosome alterations (Fig.~\ref{fig: LGG_results}A). Residual clusters 3 and 8 added further resolution: for instance, residual cluster 3 significantly modified hazard ratios within main clusters 1 and 2, while residual cluster 8 influenced outcomes among patients in main cluster 5. 
These results demonstrate that residual variation can encode information beyond that captured by dominant copy number profiles. As shown in Fig.~\ref{fig: LGG_results}B, predicted survival risk among patients aged 40-60 varied substantially across combinations of main and residual cluster memberships. Broad chromosomal alterations (main clusters) and finer-grained aberrations (residual clusters) provide complementary and additional stratification of LGG tumours. The two-stage CN-SBM framework thus uncovers biologically and clinically meaningful substructures with direct relevance for survival prediction.


\section{Discussion}

SBMs have previously been used in various biological applications: for instance, to uncover clonal structures from binary mutation matrices~\citep{myersIdentifyingTumorClones2020} and to identify rare cell populations in single-cell RNA-seq data using \textit{nested SBMs}~\citep{morelliNestedStochasticBlock2021}. The single-cell genotyper~\citep{rothClonalGenotypePopulation2016} performed clonal genotype inference via mixture modelling.
We introduced the CN-SBM, a probabilistic block model for characterizing primary copy number variation and isolating residual deviations, by jointly clustering samples and genomic bins while preserving the categorical nature of copy number states. As such, CN-SBM is well-suited to the multimodal distributions typical of copy number alterations. A key strength of the model is its use in a two-stage modelling approach: the first stage captures primary chromosomal alterations, while the second isolates structured residual variation. Our analysis shows that both components provide complementary and independently prognostic information, highlighting the clinical utility of decomposing genomic variation into coarse and fine-grained layers for survival analysis.

CN-SBM can scale effectively to large cohorts and finer bin resolutions using stochastic variational inference (see Appendix~\ref{appendix: svi}) and exhibits stable convergence across diverse datasets. In contrast, existing baselines such as Blockcluster frequently encounter convergence issues and require repeated initializations to produce stable results. While CN-SBM captures important genomic substructure, it currently assumes independence across genomic bins, which may limit its ability to detect spatially correlated patterns. For the characterization of primary variation, the CN-SBM benefits from pre-segmented input via existing copy number callers. 

Beyond clustering, CN-SBM facilitates downstream analysis through interpretable feature summaries. By leveraging the genomic ordering of bins, contiguous regions sharing cluster labels can be merged into higher-level segments, enabling the computation of summary statistics such as event frequency and length. These derived features may support the identification of mutational signatures or focal aberrations and enhance the integration of CN-SBM outputs into broader analytical workflows.

An important avenue for future work lies in extending CN-SBM to integrate multi-omic data. Recent single-cell technologies enable simultaneous DNA and RNA profiling (e.g.~\citet{macaulayGTseqParallelSequencing2015}), offering the opportunity to model joint structure across copy number alterations and gene expression. A multimodal extension of CN-SBM, incorporating shared and modality-specific latent variables, could improve stratification and yield deeper insight into the relationship between genomic alterations and transcriptional programs.

\acksection
YP and KL acknowledge the support of the CANSSI Graduate Student Enrichment Scheme (GSES). BBR acknowledges the support of NSERC: RGPIN2020-04995, RGPAS-2020-00095. YP acknowledges the support of the Canada Research Chair Tier 2 program and the NSERC Discovery Grant. We thank Charles Gadd for his insightful discussions and his assistance in accessing the TCGA datasets.

\bibliographystyle{apalike}
\bibliography{references}

\clearpage
\appendix

\section{Supplementary Material: CN-SBM}
\label{section: appendix: sbm}

\subsection{Joint Distribution and ELBO for the CN-SBM}
\label{appendix: sbm_details}


The joint distribution of the CN-SBM expands as
\begin{equation*}
\begin{split}
    p(\bfC, \bfdG, \bfdH, \bm{\pi}^g, \bm{\pi}^h,  \{\bm{\pi}^{(k,l)}\}) &=
    \underbrace{p(\bfC | \bfdG, \bfdH, \{\bm{\pi}^{(k,l)}\})}_{\text{model}}
    \cdot \underbrace{p(\bfdG | \bm{\pi}^g)}_{\text{cells}}
    \cdot \underbrace{p(\bfdH | \bm{\pi}^h)}_{\text{bins}}
    \cdot \underbrace{p(\bm{\pi}^g)}_{\text{prior}}
    \cdot \underbrace{p(\bm{\pi}^h)}_{\text{prior}}
    \cdot \underbrace{p(\{\bm{\pi}^{(k,l)}\})}_{\text{prior}} \\
    &= \underbrace{\prod_{i=1}^N \prod_{j=1}^M \pi^{(g_i,h_j)}_{c_{ij}}}_{\text{data model}}
    \cdot \underbrace{\prod_{i=1}^N \pi^g_{g_i}}_{\text{cell clusters}}
    \cdot \underbrace{\prod_{j=1}^M \pi^h_{h_j}}_{\text{bin clusters}} \\
    & \quad
    \underbrace{\frac{1}{B(\alpha)^{K\cdot L}} \prod_{k=1}^K \prod_{l=1}^L \prod_{c \in \calC} (\pi^{(k,l)}_c)^{\alpha - 1}}_{\text{block distribution prior}}
    \cdot \underbrace{\frac{1}{B(\alpha^g)^K} \prod_{k=1}^K (\pi^g_k)^{\alpha^g - 1}}_{\text{cell cluster prior}}
    \cdot \underbrace{\frac{1}{B(\alpha^h)^L} \prod_{l=1}^L (\pi^h_l)^{\alpha^h - 1}}_{\text{bin cluster prior}}.
\end{split}
\end{equation*}


The ELBO can be computed as
\begin{equation*}
\begin{split}
    \log p(\bfC)
    & \geq \bbE_{q}[\log p(\bfC, \bfdG, \bfdH, \bm{\pi}^g, \bm{\pi}^h, \{\bm{\pi}^{(k,l)}\})] 
    - \bbE_{q}[\log q(\bfdG, \bfdH, \bm{\pi}^g, \bm{\pi}^h, \{\bm{\pi}^{(k,l)}\})]
    \\
    &= \bbE_{q}[\log p(\bfC | \bfdG, \bfdH, \{\bm{\pi}^{(k,l)}\})] -\DKL (q(\bfdG, \bfdH, \bm{\pi}^g, \bm{\pi}^h, \{\bm{\pi}^{(k,l)}\}) || p(\bfdG, \bfdH, \bm{\pi}^g, \bm{\pi}^h, \{\bm{\pi}^{(k,l)}\})) \\
    &= \bbE_{q}[\log p(\bfC | \bfdG, \bfdH, \{\bm{\pi}^{(k,l)}\})] \\  
    &\quad - \DKL (q(\bfdG, \bm{\pi}^g) || p(\bfdG, \bm{\pi}^g))
    - \DKL (q(\bfdH, \bm{\pi}^h) || p(\bfdH, \bm{\pi}^h))
    - \DKL (q(\{\bm{\pi}^{(k,l)}\}) || p(\{\bm{\pi}^{(k,l)}\})),
\end{split}
\end{equation*}
where the individual terms are given by
\begin{align*}
    \bbE_{q}\left[\log p(\bfC | \bfdG, \bfdH, \{\bm{\pi}^{(k,l)}\})\right]
    &= \sum_{i,j} \bbE_q\left[\log \prod_{k,l} (\pi^{(k, l)}_{c_{ij}})^{\indicator(g_i = k)\indicator(h_j = l)}\right] \\
    &= \sum_{i,j} \sum_{k,l} \phi^g_{ik} \phi^h_{jl} \cdot \bbE_q\left[ \log (\pi^{(k, l)}_{c_{ij}})\right] \\
    &= \sum_{i,j} \sum_{k,l} \phi^g_{ik} \phi^h_{jl} \left(\psi(\gamma^{(k,l)}_{c_{ij}}) - \psi\left(\sum_{c \in \calC} \gamma^{(k,l)}_c \right) \right) \\
    \DKL (q(\bfdG, \bm{\pi}^g) || p(\bfdG, \bm{\pi}^g)) &= \sum_i \DKL(q(g_i, \bm{\pi}^g) || p(g_i, \bm{\pi}^g))\\
    &= \DKL(q(\bm{\pi}^g) || p(\bm{\pi}^g)) + \sum_i \bbE_q\left[\log \frac{q(g_i)}{p(g_i | \bm{\pi}^g)}\right]\\
    \DKL (q(\bfdH, \bm{\pi}^h) || p(\bfdH, \bm{\pi}^h)) &= \sum_j \DKL(q(h_j, \bm{\pi}^h) || p(h_j, \bm{\pi}^h))\\
    &= \DKL(q(\bm{\pi}^h) || p(\bm{\pi}^h)) + \sum_j \bbE_q\left[\log \frac{q(h_j)}{p(h_j | \bm{\pi}^h)}\right]\\
    \DKL (q(\{\bm{\pi}^{(k,l)}\}) || p(\{\bm{\pi}^{(k,l)}\})) &= \sum_{k,l} \DKL(q(\bm{\pi}^{(k,l)}) || p(\bm{\pi}^{(k,l)})).
\end{align*}
Here, the expected log-likelihood can be expanded as
\begin{align*}
    \bbE_q[\log p(g_i | \bm{\pi}^g)] &= \bbE_q[\log \pi^g_{g_i}] = \bbE_q\left[\log \prod_k (\pi^g_k)^{\indicator(g_i = k)}\right]
    = \sum_k \phi^g_{ik} \bbE_q[\log \pi^g_{k}] \\
    \bbE_q[\log p(h_j | \bm{\pi}^h)] &= \bbE_q[\log \pi^h_{h_j}] = \bbE_q\left[\log \prod_l (\pi^h_j)^{\indicator(h_j = l)}\right]
    = \sum_l \phi^h_{jl} \bbE_q[\log \pi^h_{l}] \\
\end{align*}
and $\bbE_q[\log q(g_i)] = -\sum_k \phi_{ik} \log \phi_{ik}$ is the entropy of a multinomial distribution. The KL-divergence of two Dirichlet distributions is based on the expectation of a log-transformed Dirichlet variable, i.e.
\begin{equation*}
\begin{split}
    \DKL(q(\bm{\pi}) || p(\bm{\pi})) &= \log \frac{\Gamma(\sum_k \alpha^q_k)}{\Gamma(\sum_k \alpha^p_k)} + \sum_k \log \frac{\Gamma(\alpha^p_k)}{\Gamma(\alpha^q_k)} + \sum_k(\alpha^q_k - \alpha^p_k)\bbE_q[\log \pi_k]\\
    &= \log \frac{\Gamma(\sum_k \alpha^q_k)}{\Gamma(\sum_k \alpha^p_k)} + \sum_k \log \frac{\Gamma(\alpha^p_k)}{\Gamma(\alpha^q_k)} + \sum_k(\alpha^q_k - \alpha^p_k)\left[ \psi(\alpha^q_k) - \psi\left(\sum_j \alpha^q_j\right) \right],
\end{split}
\end{equation*}
where $p(\bm{\pi}) \sim \text{Dir}_{[0, \ldots, K]}(\bm{\alpha}^p)$ and $q(\bm{\pi}) \sim \text{Dir}_{[0, \ldots, K]}(\bm{\alpha}^q)$.

\subsection{Stochastic Variational Inference}
\label{appendix: svi}

Stochastic Variational Inference (SVI)~\citep{hoffmanStochasticVariationalInference2013} extends the CAVI framework to enable scalable inference in large datasets. The core idea is that in latent variable models, each observation is typically associated with local latent variables ($g_i$, $h_j$), which influence the posterior over global variables ($\bm{\pi}^g, \bm{\pi}^h, \{\bm{\pi}^{(k,l)}\}$). While standard CAVI requires full sweeps over all local variables before updating global parameters, SVI performs stochastic optimization by sampling a mini-batch of data points instead and using these intermediate values to update the global parameters, reducing the per-iteration cost. Since the underlying variational updates remain structurally identical to the ones in Algorithm~\ref{algo:sbm_cavi}, adapting our model to SVI is straightforward. The resulting SVI procedure is outlined in Algorithm~\ref{algo:sbm_svi}.

\begin{algorithm}[!htb]
\caption{Stochastic variational inference for the CN-SBM (Stochastic VI)}
\begin{algorithmic}
\Require Copy number matrix $\bfC = (c_{ij})$, learning schedule $(\eta_t)_{t \geq 1}$.
\State Set $t=1$. Initialize local parameters $\bm{\phi}^g, \bm{\phi}^h$ and global parameters $\bm{\gamma}^g, \bm{\gamma}^h, \{\bm{\gamma}^{(k,l)}\}$.
\While{ELBO not converged}
    \State \textbf{Local updates:} Sample subset $\calI_N \subseteq [N], \calI_M \subseteq [M]$
    \begin{itemize}
        \item for row $i \in \calI_N$ and row clusters $k = 1, \ldots, K$
        \begin{itemize}
            \item $q(g_i = k) \propto \exp\left(\sum_{j=1}^M \sum_{l=1}^L \phi^h_{jl} \cdot \left(\psi(\gamma^{(k,l)}_{c_{ij}}) - \psi\left(\sum_{c \in \calC} \gamma^{(k,l)}_c \right) \right)  + \psi(\gamma^g_k) - \psi\left(\sum_{k=1}^K \gamma^g_k\right)  \right)$
        \end{itemize}
        \item for col $j \in \calI_M$ and col clusters $l = 1, \ldots, L$
        \begin{itemize}
            \item $q(h_j = l) \propto \exp\left(\sum_{i=1}^N \sum_{k=1}^K \phi^g_{ik} \cdot \left(\psi(\gamma^{(k,l)}_{c_{ij}}) - \psi\left(\sum_{c \in \calC} \gamma^{(k,l)}_c \right) \right)  + \psi(\gamma^h_l) - \psi\left(\sum_{l=1}^L \gamma^h_l\right)  \right)$
        \end{itemize}
    \end{itemize}
    \State \textbf{Intermediate global updates for cluster and cluster proportions}
    \begin{itemize}
        \item $\bm{\hat\gamma}^g = \alpha^g + \frac{N}{|\calI_N|}\sum_{i \in \calI_N} \phi^g_{i, \cdot}$
        \item $\bm{\hat\gamma}^h = \alpha^h + \frac{M}{|\calI_M|}\sum_{j \in \calI_M} \phi^h_{j, \cdot}$
        \item $\bm{\hat\gamma}^{(k,l)} = \alpha + \frac{NM}{|\calI_N|\cdot |\calI_M|}\sum_{i \in \calI_N}\sum_{j \in \calI_M}  \phi^g_{ik} \phi^h_{jl} \indicator(c_{ij}=\cdot), \quad \quad 1 \leq k \leq K, 1\leq l \leq L.$
    \end{itemize}
    \State \textbf{Update global estimates}
    \begin{itemize}
        \item $\bm{\gamma}^g \leftarrow (1-\eta_t)\bm{\gamma}^g + \eta_t \bm{\hat\gamma}^g$
        \item $\bm{\gamma}^h \leftarrow (1-\eta_t)\bm{\gamma}^h + \eta_t \bm{\hat\gamma}^h$
        \item $\bm{\gamma}^{(k,l)} \leftarrow (1-\eta_t)\bm{\gamma}^{(k,l)} + \eta_t \bm{\hat\gamma}^{(k,l)}$
    \end{itemize}
    \State $t \gets t+1$
\EndWhile
\end{algorithmic}
\label{algo:sbm_svi}
\end{algorithm}


\subsection{Importance Weighting for the CN-SBM with Missing Data}
\label{section: missing}

In section~\ref{section: MNSBM}, we performed model inference by maximizing the ELBO for the observed data log-likelihood, $\log p_{\theta}(\bfC_\text{obs})$, effectively ignoring the missing data. However, a more principled approach may be to optimize the ELBO for the full data-generating process, i.e., bounding $\log p(\bfC_\text{obs}, \bfC_\text{mis}) = \log p(\bfC)$ instead. This ensures that the latent variables recover structure across the entire dataset, rather than just the observed portion.
Leting $\bfB = (b_{ij})$ denote the binary missingness matrix as before the ELBO for the full data is given by
\begin{equation*}
    \log p(\bfC)
    \geq \bbE_{q}[\log p(\bfC | \bfdG, \bfdH, \{\bm{\pi}^{(k,l)}\})] -\DKL (q(\bfdG, \bfdH, \bm{\pi}^g, \bm{\pi}^h, \{\bm{\pi}^{(k,l)}\}) || p(\bfdG, \bfdH, \bm{\pi}^g, \bm{\pi}^h, \{\bm{\pi}^{(k,l)}\})).
\end{equation*}
In the presence of missing data, we can approximate the likelihood term using inverse propensity weights (IPW):
\begin{equation*}
\begin{split}
    \bbE_{q}[\log p(\bfC | \bfdG, \bfdH, \{\bm{\pi}^{(k,l)}\})] &= \sum_{i=1}^N \sum_{j=1}^M \bbE_{(\bfdZ \sim q)}[\log p(c_{ij} | \bfdG, \bfdH, \{\bm{\pi}^{(k,l)}\})] \\
    &= \sum_{i=1}^N \sum_{j=1}^M \bbE\left[\frac{B_{ij}}{\zeta_{ij}}\bbE_{(\bfdZ \sim q)}[\log p(c_{ij} | \bfdG, \bfdH, \{\bm{\pi}^{(k,l)}\})] \right] \\
    &\approx \sum_{(i,j) \in \calE^\text{obs}} \frac{1}{\zeta_{ij}}\bbE_{(\bfdZ \sim q)}[\log p(c_{ij} | \bfdG, \bfdH, \{\bm{\pi}^{(k,l)}\})] \\
    &\approx \sum_{(i,j) \in \calE^\text{obs}} \frac{1}{\hat{\zeta}_{ij}}\bbE_{(\bfdZ \sim q)}[\log p(c_{ij} | \bfdG, \bfdH, \{\bm{\pi}^{(k,l)}\})]
\end{split}
\label{eq: sbm_elbo_ipw}
\end{equation*}
Here $\calE^\text{obs}$ is the set of observed indices and $\zeta_{ij}$ denotes the probability of observing the copy number $c_{ij}$, which may depend on observed data or additional covariates. This term can be computed using only the observed data, and therefore does not require inference over the missing entries. We make two approximations: we approximate the expectation using a single observed sample by setting $b_{ij} = 1$ and estimate the observation propensity as $\hat{\zeta}_{ij}$. The variance of this estimator can become large when individual propensity scores are low. To mitigate this, it may be beneficial to exclude columns with a high proportion of missing data. We plan to perform a sensitivity analysis to understand the impact of missing data in the various missingness scenarios.

This approach relies on specifying an appropriate model for the observation mechanism and how the missing data is generated.
In the Missing at Random (MAR) setting, the observation propensity can initially be modeled using logistic regression, depending only on the row and column indices. Alternatively, a simple heuristic based on observation frequencies can be used: the propensity could be approximated as $\zeta_{ij} = \frac{|\calI_{i,\text{obs}}|}{M}\frac{|\calJ_{j,\text{obs}}|}{N}$, where $|\calI_{i,\text{obs}}|$ and $|\calJ_{j,\text{obs}}|$ denote the number of observed entries in row i and column j, respectively. This approach assumes that missingness is independent across rows and columns and provides a simple, data-driven estimate of the observation probability.

\paragraph{Weighted CAVI Updates}

When using importance weights for the CN-SBM, we can also apply approximate update steps similar to the CAVI update formulations in Algorithm~\ref{algo:sbm_cavi}:
\begin{equation}
\begin{split}
    q(g_i = k| \bm{\phi}^g) 
    &\propto \exp\left(\sum_{j=1}^M \sum_{l=1}^L \bbE \left[\log \left(\pi^{(k,l)}_{c_{ij}}\right)^{\indicator(h_j = l)}\right] + \bbE[\log \pi^g_k]  \right) \\
    &\approx \exp\left(\sum_{j: (i,j) \in \calE^\text{obs}} \sum_{l=1}^L \frac{1}{\zeta_{ij}} \bbE[\indicator(h_j = l)] \bbE[\log \pi^{(k,l)}_{c_{ij}}] + \bbE[\log \pi^g_k] \right) \\
    q(h_j = l| \bm{\phi}^h)
    &\propto \exp\left(\sum_{i=1}^N \sum_{k=1}^K \bbE \left[\log \left(\pi^{(k,l)}_{c_{ij}}\right)^{\indicator(g_i = k)]}\right] + \bbE[\log \pi^h_l] \right) \\
    &\approx \exp\left(\sum_{i: (i,j) \in \calE^\text{obs}} \sum_{k=1}^K \frac{1}{\zeta_{ij}} \bbE[\indicator(g_i = k)] \bbE[\log \pi^{(k,l)}_{c_{ij}}] + \bbE[\log \pi^h_l] \right) \\
    \bm{\gamma}^{(k,l)} &= \alpha + \sum_{i=1}^N\sum_{j=1}^M  \bbE[\indicator(g_i = k)\indicator(h_j = l)] \indicator(c_{ij}=\cdot) \\
    &\approx \alpha + \sum_{(i,j) \in \calE^\text{obs}} \frac{1}{\zeta_{ij}} \bbE[\indicator(g_i = k)\indicator(h_j = l)] \indicator(c_{ij}=\cdot),
    \quad \quad 1 \leq k \leq K, 1\leq l \leq L.
\end{split}
\end{equation}

The other two update steps for $\bm{\gamma}^g$ and $\bm{\gamma}^h$ remain exact as they do not depend on the observed data.
However, because the importance-weighted updates are approximations, we no longer have a guarantee that each iteration will increase the ELBO. Therefore, we consider the algorithm to have converged when a moving average of the ELBO shows no significant change over time.

\clearpage
\section{Supplementary Material: Benchmarks and Results}

\subsection{Implementation and Data Sources}

\paragraph{Implementation Details} 

The CN-SBM model was implemented in \texttt{Python 3.9} using \texttt{JAX}, which enables efficient linear algebra via just-in-time (JIT) compilation. Training for the CN-SBM was performed on a virtual machine equipped with an NVIDIA RTX 3080 GPU (10 GB memory, using a single GPU core). The CAVI algorithm terminates when the ELBO improvement falls below a threshold of $10^{-4}$. For datasets at 500kb resolution with 2,500–5,000 cells, convergence typically occurs within 20–30 minutes, with most of the time spent on JIT compilation and initialization. For higher resolutions (e.g. 100kb or 50kb) or larger cell counts, we recommend stochastic variational inference to avoid memory overruns.

\begin{figure}[!htb]
    \centering
    \includegraphics[width = 1\linewidth]{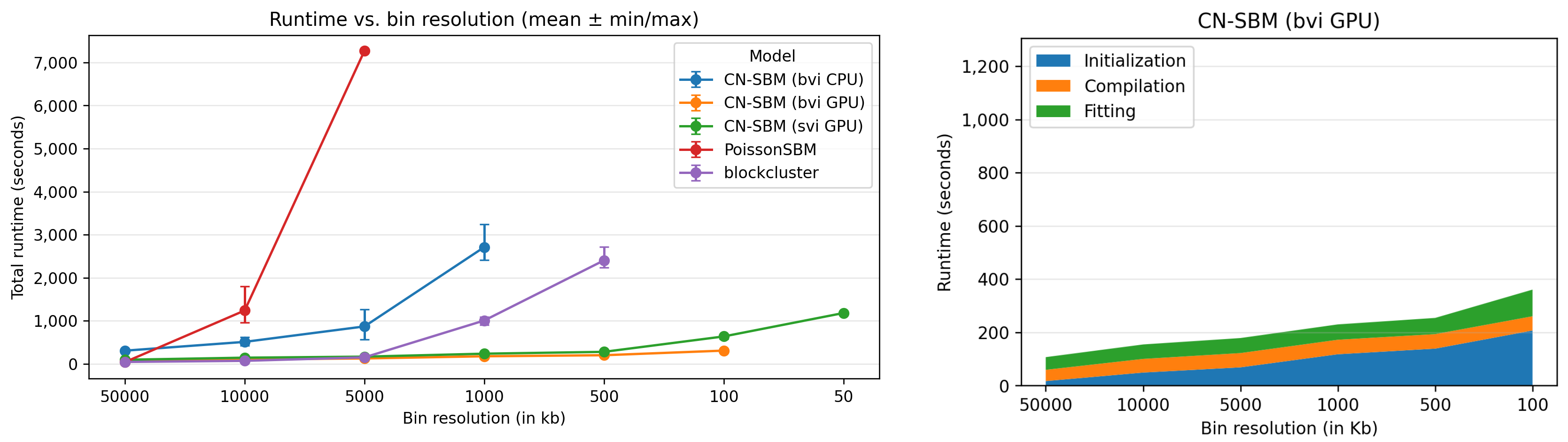}
    \caption{\textbf{Left:} Runtimes for the CN-SBM, PoissonSBM, and Blockcluster on the same machine. \texttt{bvi} refers to Algorithm~\ref{algo:sbm_cavi} and \texttt{svi} to Algorithm~\ref{algo:sbm_svi}. \textbf{Right:} Runtime decomposition of the CN-SBM into spectral bi-clustering initialization, JAX model compilation, and CAVI model fitting.}
    \label{fig: runtimes}
\end{figure}

Runtimes for the TCGA BRCA dataset ($N=998$) at varying resolutions are provided in Fig.~\ref{fig: runtimes}, where we also compared the runtimes for the PoissonSBM and the Blockcluster model on the same machine and demonstrate the scalability of the algorithm to finer resolution data using stochastic variational inference.

We use the ICL as a measure of model fit on the complete dataset. We approximate it as
\begin{equation*}
    \text{ICL}(\bfC; \hat{\bfdG}_\text{MAP}, \hat{\bfdH}_\text{MAP}) \approx
    \log p(\bfC, \hat{\bfdG}_\text{MAP}, \hat{\bfdH}_\text{MAP} | \hat{\bm{\pi}}^g, \hat{\bm{\pi}}^h, \{\hat{\bm{\pi}}^{(k,l)}\}) - \text{pen}_\text{ICL}(\bfC, \hat{\bfdG}_\text{MAP}, \hat{\bfdH}_\text{MAP})
\label{eq: ICL}
\end{equation*}
where the penalty term is defined as
\begin{equation*}
    \text{pen}_\text{ICL}(\bfC, \hat{\bfdG}_\text{MAP}, \hat{\bfdH}_\text{MAP}) =
    - \frac{1}{2} \left[
        \underbrace{(K-1)\log(N)}_{\text{Penalty on }\hat{\bfdG}_\text{MAP}}
        + \underbrace{(L-1)\log(M)}_{\text{Penalty on }\hat{\bfdH}_\text{MAP}}
        + \underbrace{(n_\text{cat} - 1) \cdot K \cdot L}_{\text{Block Complexity}} \cdot \underbrace{\log |\calE^\text{obs}|}_{\text{Data Size}}
    \right],
\end{equation*}
Here, $|\calE^\text{obs}|$ is the number of observed data points in the copy number matrix, and $\hat{\bm{\pi}}^g, \hat{\bm{\pi}}^h, \{\hat{\bm{\pi}}^{(k,l)}\}$ are maximum likelihood estimates. ICL offers a model-based assessment that balances goodness-of-fit with model complexity, by combining a likelihood term involving the maximum a posteriori (MAP) estimate with a penalty accounting for the number of effective clusters. This enables model comparison based on the final clustering configuration inferred from the MAP estimate.

\paragraph{Data Sources} The CNAsim package is available at \href{https://github.com/samsonweiner/CNAsim/tree/main}{https://github.com/samsonweiner/CNAsim/tree/main}. We used the command \texttt{cnasim -m 0 -n 2500 -c 7 -v -U -B 500000 -E1 0.04 -E2 0.1 -P 24} to generate 2,500 single-cell copy number profiles. TCGA data with ASCAT-inferred copy numbers were preprocessed in the repository at \href{https://github.com/cwlgadd/TCGA}{https://github.com/cwlgadd/TCGA}
using the \texttt{TCGA.data\_modules.ascat.loaders.LoadASCAT()} function. Patient survival data can be extracted using the \texttt{RTCGA} package, see \href{https://rtcga.github.io/RTCGA/reference/survivalTCGA.html}{https://rtcga.github.io/RTCGA/reference/survivalTCGA.html} for additional information. Single-cell data from~\citet{funnellSinglecellGenomicVariation2022} can be obtained online at Zenodo: \href{https://zenodo.org/records/6998936}{https://zenodo.org/records/6998936}.


\subsection{Additional Benchmark Details} 
\label{section: add_benchmark_details}

We initially included spectral co-clustering (SpecCo) as a baseline method, however, it consistently performed worse than all other models, which is why we excluded this method from the model comparisons. Table~\ref{tab:combined_entropy} presents supplementary perfomrance metrics: classification accuracy on held-out copy numbers and the average normalized entropy of clusters for models trained on the complete dataset. We considered accuracy as a secondary metric as it involves two approximations: converting soft clusters to hard assignments and selecting a MAP category from the categorical block distributions. Models are not explicitly optimized for low entropy, so this serves as a sanity check on cluster quality.

\begin{table}[ht]
\centering
\small
\vspace{1mm}
\renewcommand{\arraystretch}{1.1}
\setlength{\tabcolsep}{2.5pt}
\begin{tabular}{|c|l|c|ccc|ccc|}
\hline
 & \textbf{Method} & \textbf{CNAsim} & \multicolumn{3}{c|}{\textbf{TCGA}} & \multicolumn{3}{c|}{\textbf{Funnell}} \\
\cline{3-9}
 &  & \makecell{Simulated \\ $(N=2500)$} & \makecell{BRCA \\ $(N=998)$} & \makecell{OV \\ $(N=532)$} & \makecell{LGG \\ $(N=490)$} & \makecell{OV2295 \\ $(N=1084)$} & \makecell{SA1096 \\ $(N=802)$} & \makecell{SA535 \\ $(N=1801)$} \\
\hline\hline
\multirow{6}{*}{\makebox[6pt][c]{\rotatebox[origin=c]{90}{\textbf{Acc.} $\times 10^{2}$ ($\uparrow$)}}}
& SpecBi (bist)         & 82.26 ± 0.05  & 52.33 ± 0.03 & 42.71 ± 0.42 & 79.60 ± 0.56 & 70.05 ± 0.94 & 60.89 ± 0.67 & 62.97 ± 1.14 \\
& SpecBi (log)          & 82.48 ± 0.06  & 52.77 ± 0.19 & 43.85 ± 0.24 & 80.38 ± 0.32 & 69.22 ± 1.15 & 61.83 ± 0.77 & 66.75 ± 0.87 \\
& KMeans                & 93.36 ± 0.20  & 62.19 ± 0.24 & 51.05 ± 0.15 & \textbf{87.77 ± 0.16} & \textbf{81.52 ± 0.39} & \textbf{81.32 ± 0.45} & \textbf{88.41 ± 0.43} \\
& PoissonSBM            & 89.57 ± 3.77  & 14.63 ± 0.41 & 13.31 ± 0.79    & 21.32 ± 0.63 & 60.91 ± 1.51 & 73.81 ± 0.70 & 88.19 ± 0.76 \\
& Blockcluster                 & 93.29 ± 0.22  & 62.66 ± 0.24 & 52.35 ± 0.52    & 86.38 ± 0.74 & 79.69 ± 0.60 & 80.36 ± 0.40 & 85.43 ± 1.08 \\
\cline{2-9}
& CN-SBM (ref.)         & \textbf{93.81 ± 0.05} & \textbf{63.12 ± 0.09} & \textbf{53.01 ± 0.45} & 87.42 ± 0.27 & 80.99 ± 0.89 & 81.23 ± 0.62 & 87.43 ± 0.31 \\
\hline\hline
\multirow{6}{*}{\makebox[6pt][c]{\rotatebox[origin=c]{90}{\textbf{Entr.} $\times 10^{-2}$ ($\downarrow$)}}}
  & SpecBi (bist)         & 24.6 ± 0.0       & 51.4 ± 0.2   & 58.5 ± 0.3   & 26.2 ± 0.3     & 33.9 ± 0.4     & 40.7 ± 0.8   & 41.2 ± 2.3   \\
 & SpecBi (log)          & 24.5 ± 0.2       & 51.4 ± 0.3   & 57.1 ± 0.2   & 26.3 ± 0.3     & 33.7 ± 0.8     & 40.3 ± 0.5   & 36.8 ± 1.5   \\
 & KMeans                & 11.1 ± 0.3       & 41.1 ± 0.2   & 48.6 ± 0.1   & 15.2 ± 0.2     & 22.3 ± 0.2     & 22.3 ± 0.2   & 16.0 ± 0.2   \\
 & PoissonSBM            & 10.5 ± 0.0       & 38.7 ± 0.0   & \textbf{47.3 ± 0.0}   & \textbf{14.7 ± 0.0}     & 22.4 ± 0.0     & \textbf{21.3 ± 0.0}  & 15.7 ± 0.0   \\
 & Blockcluster                 & 10.4 ± 0.2       & 39.0 ± 0.1   & 47.6 ± 0.1   & 16.3 ± 0.1     & 24.7 ± 0.7     & --   & 17.0 ± 0.1   \\
\cline{2-9}
 & CN-SBM (ref.)         &  \textbf{9.9 ± 0.0}       & \textbf{38.5 ± 0.1}   & 47.4 ± 0.1   & 15.3 ± 0.1     & \textbf{21.3 ± 0.6}     & 21.7 ± 0.2   & \textbf{15.3 ± 0.3}   \\
\hline
\end{tabular}
\caption{Accuracy (top) on held-out data, normalized entropy (bottom) (mean ± std) across simulated and real datasets. Values were scaled by $10^{-2}$ for accuracy and entropy. Dashes for the entropy indicate missing values due to failed runs}
\label{tab:combined_entropy}
\end{table}

Table~\ref{tab:combined_loglik_ICL_alt} reports performance metrics (held-out loglikelihood, accuracy) based on the block probability distributions inferred from the PoissonSBM and Blockcluster implementations, rather than empirical block distributions. Since PoissonSBM uses a Poisson distribution, likelihood values are expected to be much lower as it cannot capture dominant copy number categories, e.g. $\text{Poisson}(x=2;\lambda=2) \approx 0.27$. Similarly to the results in Table~\ref{tab:combined_loglik_ICL}, PoissonSBM also shows low ICL variance, as it is trained on the full dataset using exhaustive search for cluster partitions.

For the Blockcluster implementation~\citep{bhatiaBlockclusterPackageModelBased2017}, we initially considered using the semi-supervised implementation of Blockcluster (\texttt{R-blockcluster} package) by using spectral clustering informed inputs, however, the model runs failed multiple times. As the method already employs a smart initialization, we used the default settings. Due to issues computing log-likelihoods from its probability outputs, which led to extreme values, we relied on empirical block distributions in the main results and report package-specific metrics in Table~\ref{tab:combined_loglik_ICL_alt}.

\begin{table}[ht]
\small
\centering
\vspace{1mm}
\renewcommand{\arraystretch}{1.1}
\setlength{\tabcolsep}{2.5pt}
\begin{tabular}{|c|l|c|ccc|ccc|}
\hline
 & \textbf{Method} & \textbf{CNAsim} & \multicolumn{3}{c|}{\textbf{TCGA}} & \multicolumn{3}{c|}{\textbf{Funnell}} \\
\cline{3-9}
 &  & \makecell{Simulated \\ $(N=2500)$} & \makecell{BRCA \\ $(N=998)$} & \makecell{OV \\ $(N=532)$} & \makecell{LGG \\ $(N=490)$} & \makecell{OV2295 \\ $(N=1084)$} & \makecell{SA1096 \\ $(N=802)$} & \makecell{SA535 \\ $(N=1801)$} \\
\hline\hline
\multirow{2}{*}{\makebox[6pt][c]{\rotatebox[origin=c]{90}{\textbf{LL}}}}
  & PoissonSBM            & -198.4 ± 0.1 & -209.7 ± 0.6 & -90.0 ± 0.2 & -45.3 ± 0.3 & -134.6 ± 0.8 & -163.5 ± 1.6 & -186.6 ± 0.1 \\
 & Blockcluster          & -38.2 ± 0.4 & -101.5 ± 53.5 & -66.1 ± 41.3 & -118.2 ± 92.3 & -193.0 ± 308.3 & -161.6 ± 231.9 & -181.0 ± 231.3 \\ 
\hline\hline
\multirow{2}{*}{\makebox[6pt][c]{\rotatebox[origin=c]{90}{\textbf{ICL}}}}
 & PoissonSBM            & -18.88 ± 0.00 & -9.06 ± 0.00 & -4.93 ± 0.00 & -4.05 ± 0.00 & -9.93 ± 0.00 & -7.46 ± 0.00 & -16.11 ± 0.00 \\
 & Blockcluster          & -21.98 ± 21.04 & -14.14 ± 7.91 & -3.74 ± 0.01 & -1.21 ± 0.01 & -25.95 ± 37.83 & -- & -68.42 ± 55.37 \\
\hline\hline
\multirow{2}{*}{\makebox[6pt][c]{\rotatebox[origin=c]{90}{\textbf{Acc.}}}}
 & PoissonSBM            & 93.62 ± 0.09 & 60.6 ± 0.21 & 50.53 ± 0.05 & 89.51 ± 0.28 & 80.39 ± 0.33 & 81.65 ± 0.22 & 88.59 ± 0.24\\
 & Blockcluster          & 93.29 ± 0.22 & 46.4 ± 20.11 & 37.29 ± 21.75 & 48.75 ± 37.57 & 65.81 ± 28.27 & 58.87 ± 37.46 & 67.16 ± 31.6\\ 
\hline
\end{tabular}
\caption{Held‐out log‐likelihood (top), ICL (middle), and accuracy (bottom) across simulated and real datasets using the probability distributions provided by each model implementation. Values were scaled by $10^3$ for log‐likelihood, $10^6$ for ICL, and $10^{-2}$ for the accuracy. Dashes indicate missing values due to failed runs.}
\label{tab:combined_loglik_ICL_alt}
\end{table}

Table~\ref{tab:model_runs} reports the number of successful model runs across five random seeds for each experiment. PoissonSBM completed only 3 runs on the TCGA-OV dataset due to time constraints, as it explores a large number of cluster combinations. Blockcluster failed on multiple runs, likely due to convergence issues in its package implementation.

\begin{table}[ht]
\small
\centering
\vspace{1mm}
\renewcommand{\arraystretch}{1.1}
\setlength{\tabcolsep}{3pt}
\begin{tabular}{|l|c|c|ccc|ccc|}
\hline
\textbf{Method} & \textbf{Missing} & \textbf{CNAsim} &
\multicolumn{3}{c|}{\textbf{TCGA}} &
\multicolumn{3}{c|}{\textbf{Funnell}} \\
\cline{3-9}
 &  & \makecell{Simulated\\ $(N=2500)$} &
 \makecell{BRCA\\ $(N=998)$} &
 \makecell{OV\\ $(N=532)$} &
 \makecell{LGG\\ $(N=490)$} &
 \makecell{OV2295\\ $(N=1084)$} &
 \makecell{SA1096\\ $(N=802)$} &
 \makecell{SA535\\ $(N=1801)$} \\
\hline\hline
\multirow{2}{*}{PoissonSBM}         & False & 5 & 5 & 4 & 5 & 5 & 5 & 5 \\
                              & True  & 5 & 5 & 3 & 5 & 5 & 5 & 5 \\
\hline
\multirow{2}{*}{Blockcluster} & False & 4 & 4 & 4 & 2 & 3 & 0 & 3 \\
                              & True  & 3 & 4 & 2 & 3 & 4 & 3 & 3 \\
\hline
\end{tabular}
\caption{Number of successful model runs across 5 seeds for PoissonSBM and Blockcluster.}
\label{tab:model_runs}
\end{table}

\subsection{TCGA-LGG Survival Modelling}
\label{appendix: cox_model_selection}

\paragraph{Model Selection Details} 

We observed that age, when treated as a continuous covariate, was generally not statistically significant, as indicated by its associated $p$-values. In contrast, the discretization of age into age groups more effectively captured the risk patterns present in the survival data, leading to the baseline model to include age groups.

We began by analyzing a Cox model incorporating the main variation clusters (Fig.~\ref{fig: LGG_results}A). All clusters 0-5 except 1 were found to be significant, where the $p$-value was $0.16$ for the latter. Combining cluster 1 with any of the remaining clusters would yield an overall significant model, however, merging clusters 1 and 2 was the most appropriate: these two clusters displayed highly similar genomic profiles, only differing in the copy numbers in the \texttt{X} chromosome, likely attributable to patient sex. To validate this, we refitted the CN-SBM model on the TCGA-LGG dataset excluding the \texttt{X} chromosome. In this output, samples from clusters 1 and 2 were grouped into a single, larger cluster, supporting the decision to merge them. As a result, we arrived at the model involving age groups and the main variation clusters, where clusters 1 and 2 were merged.

To assess the inclusion of residual clusters, we extended the baseline model to incorporate them (without main clusters). Residual clusters 2, 3, 4, 6, 8, and 9 were found to be significant, with clusters 3 and 8 exhibiting the largest hazard ratios, namely 6.2 (95\% CI: 2.7-14.0) and 5.9 (95\% CI: 2.0-17.6), respectively. We then explored individually stratifying the main clusters based on their residual cluster groupings, guided by the Sankey diagram (Fig.~\ref{fig: LGG_results}C). This analysis led to the final model shown in Fig.~\ref{fig: LGG_results}D, where we found that residual cluster 3 modified the effect of the merged main clusters 1 \& 2, and residual cluster 8 similarly influenced main cluster 5. In this model, we also included the interaction between main cluster 3 and residual cluster 3 (where there was no significant effect) to illustrate how the impact of residual variation can depend on the main cluster.

This analysis highlights that residual variation does not necessarily translate to significant survival effects and may interact with other unmodeled covariates. Our model selection process was designed to evaluate the contributions of both main and residual variation clusters to patient survival. We acknowledge that more comprehensive models that include additional covariates beyond age groups and copy number variation may further improve predictive and explanatory performance.

\end{document}